\documentclass[runningheads]{llncs}

\usepackage{times}
\usepackage{graphicx}
\usepackage{latexsym}
\usepackage{amsmath}
\usepackage{subfigure}
\usepackage{paralist}
\begin{document}

\title{Comparing Knowledge-Based Reinforcement Learning \\ 
       to Neural Networks in a Strategy Game}
%Guidelines for Preparing a Paper for the \\
%European Conference on Artificial Intelligence}

\author{Liudmyla Nechepurenko\inst{1} \and Viktor Voss\inst{1} \and Vyacheslav Gritsenko\inst{1}}

\authorrunning{L. Nechepurenko et al.}

\institute{Arago GmbH, Eschersheimer Landstr. 526-532, 60433 Frankfurt am Main, Gernamy \email{\{lnechepurenko,vvoss,vgritsenko\}@arago.co}}

\maketitle
\bibliographystyle{splncs04}

\begin{abstract}
  The paper reports on an experiment, in which a Knowledge-Based Reinforcement Learning (KB-RL) method was compared to a Neural Network (NN) approach in solving a classical Artificial Intelligence (AI) task. In contrast to NNs, which require a substantial amount of data to learn a good policy, the KB-RL method seeks to encode human knowledge into the solution, considerably reducing the amount of data needed for a good policy. By means of Reinforcement Learning (RL), KB-RL learns to optimize the model and improves the output of the system. Furthermore, KB-RL offers the advantage of a clear explanation of the taken decisions as well as transparent reasoning behind the solution. 

  The goal of the reported experiment was to examine the performance of the KB-RL method in contrast to the Neural Network and to explore the capabilities of KB-RL to deliver a strong solution for the AI tasks. The results show that, within the designed settings, KB-RL outperformed the NN, and was able to learn a better policy from the available amount of data. These results support the opinion that Artificial Intelligence can benefit from the discovery and study of alternative approaches, potentially extending the frontiers of AI.
\end{abstract}

\section{INTRODUCTION}

Machine Learning (ML) is one of the most recent trends in Artificial Intelligence. To a high extend, ML owns its popularity to the development of Neural Networks (NNs) and, particularly, Deep Neural Networks. The fact that NNs can find solutions for previously unsolvable tasks, drew a lot of attention to NNs and ML, while decreasing the attention to other approaches.
At the same time, NNs face a number of challenges that can result in the lower performance of the NN method \cite{disadvantagesNNOverview}. For example, NNs are data greedy \cite{disadvantagesNNData}, they require an extensive amount of data in order to train an effective model. 
One of the crucial hardships employing NNs is their unexplained output \cite{disadvantagesNNBlackBox}: working as a 'back box', NNs cannot justify their output and, thus, lack to be trusted \cite{disadvantagesNNExplain}.

Studying new approaches opens Artificial Intelligence to new possibilities and facilitates a more informed choice of the technology when it comes to finding a suitable method for a research problem. Therefore, we believe that this report positively contributes to the healthy dynamic of the AI field by encouraging the diversity of its methodology and in-depth analysis of various methods. The objective of this article is to study the KB-RL method and compare it to Neural Networks in order to bring more light on the method's performance, its characteristics and the benefits of the method for a corresponding problem area. 

In this report, KB-RL conforms to the idea that exploiting human knowledge can shorten the time and data needed for machines to learn. KB-RL consists of two techniques: the Knowledge-Based technique and Reinforcement Learning. In the Knowledge-Based approach \cite{AkerkarKBSbook}, the knowledge is the essence of the system, it expresses the facts, information and skills needed to automate a particular task or a problem. 

In KB-RL, RL is added to the Knowledge-Based System (KBS) as a conflict resolution strategy targeted at integrating the alternative knowledge, or multi-expert knowledge, into one knowledge base.

The core difference between KB-RL and the traditional KBSs is their approach to knowledge. Classical KBSs strictly require the knowledge to be unambiguous. This means, if there is more than one rule to be matched to the current situation, the inference engine cannot proceed and requires conflict resolution. The conflict resolution is based on the assumption that only one rule can be correct in the given context, while others are attributed to mistakes \cite{Khalil}. In contrast, KB-RL considers conflicting rules as possible variations of the solution and employs RL to learn which variant delivers the best outcome for the defined task. Therefore, KB-RL easily accommodates ambiguous knowledge seeing it as knowledge of multiple experts that have different opinions or different strategies for the problem solution.
Report \cite{firstPaper} illustrates this principle on the example of the multi-strategy game CIVILIZATION. The report shows that, based on the several available strategies to play the game, the system learned to improve the gameplay, increasing the winning rate for the trained agent. 

Following the work \cite{firstPaper}, we chose FreeCiv as a benchmark for the reported comparison. This allowed us to reuse the open-sourced framework of FreeCiv, the knowledge base and the data that was openly published at \cite{freecivGames1}.
Due to the highly complex nature of FreeCiv, we addressed only a separate sub-task of the game. In particular, we chose to optimize the selection of the cities locations that would lead the nation to the maximal generated natural resources. Several reasons influenced the choice of the task. Besides being of reasonable complexity, this task involves image analysis - the area where NNs have shown a great performance. Moreover, there can be various strategies to approach settlement in the game, which is an appropriate setting for the KB-RL approach. Generating natural resources is a critical activity in the game due to natural resources being the main component for a nation to further develop. 
City locations are closely linked to the amount of generated natural resources because natural resources are generated by the city from the tiles within the city borders, and their amount depends, in the first place, on the map tiles properties. Therefore, wisely choosing the city region acts as a fundamental prerequisite for generating rich natural resources.

For the purpose of this work, the selected task was solved twice, with the KB-RL and the Neural Networks approach. The difference in their performance became a determining factor for the analysis of this experiment. The results of this work satisfy the goal to provide better understanding of KB-RL's performance and capabilities, and the advantages it can give in contrast to the Neural Networks. It is important to note that the experiment could have other design decisions that may be questioned by the reader throughout the article. Unfortunately, we were limited by the time dedicated to this project to explore all possible designs.

\section{RELATED WORK}
Comparative studies are commonly used to investigate methods and to gain better understanding of their strengths and weaknesses. Being one of the most popular AI techniques nowadays, NNs have been compared to other approaches in numerous studies, for example \cite{compstudy2,compstudy4,compstudy1}.

Previously, there were many instances of combining the Knowledge-Based approach with Machine Learning techniques into one method, for example \cite{combiningKBML,combiningKBANN,combiningKBMLOnco,ILP}. For several decades, a large number of studies tackled the idea of combining these two methods from various angles. To the best of our knowledge, applying RL as a conflict resolution strategy in a multi-expert knowledge-based system for incremental solution improvement was first proposed in \cite{firstPaper}. As the authors of \cite{firstPaper} suggest, it is difficult to relate to similar studies that combine the Knowledge-Based and the Machine Learning approaches due to the diversity of the addressed problems by these studies. Therefore, it was not our objective to review all the methods combining Knowledge-Based approaches and Machine Learning in the related work section. We believe that KB-RL, as well as other methods, are worth considering and can deliver comprehensive results in numerous applications. Our primary goal was to examine the KB-RL method, for which a comparative study was regarded as an effective mechanism.

Games are a recognized benchmark for testing AI algorithms. Strategy games, such as CIVILIZATION, present an attractive case for AI research due to their high complexity. To be aware of the previous applications, we reviewed several reports on employing FreeCiv as a proving point for AI algorithms. Most studies addressed specific elements of the game, with only a few works playing the entire game. In 2004, A. Houk used a symbolic approach and reasoning to develop an agent for playing FreeCiv \cite{AgentForFreeciv}. 
A series of articles explored Knowledge-Based related approaches for tasks, such as building and defending cities and population growth, see \cite{Jones2009MetareasoningFA,jones2009,jones2005,jones2005_2,Hinrichs2,Hinrichs1}.
In 2014, Branavan et al. employed Natural Language Processing (NLP) to improve the player performance in CIVILISATION II \cite{WinByReadingManuals}. 
In 2009, S. Wender, implemented several modifications of Sarsa and Q-learning algorithms to learn the potentially best city sites \cite{SettlementFreeciv}. 
References \cite{GAandFreeciv2} and \cite{GAandFreeciv1} explored the utilization of Genetic Algorithms for the optimization of city placement and city development in FreeCiv.
Overall, all the reviewed reports showed improved results of using the proposed methodology. 

Our work differs from the reviewed methods in a number of settings, such as methods, episode length, complexity of the game configurations and the selected metric. The next sections give a detailed explanation of the performed experiment.

\section{TASK DEFINITION} \label{SelectedParameters}
The amount of natural resources generated in the FreeCiv game is implemented through the points of different types that are produced by cities in every game turn. We call the amount of generated natural resource from all cities in one game: the \textbf{total game output} (TGO). We aimed to maximize the total game output by analysing the map, predicting the quality of map clusters (regions of 5x5 without corner tiles as it is shown in Figure \ref{fig:mapcluster}, which are 21 tiles) with respect to the TGO and building the cities in the places with the highest predicted quality.

Cities generate natural resources from those tiles within city borders that have working citizen on the tile. City borders may reach terrain within the 5x5 region centered on the city, minus its corners. To extract resources from a tile, the player must have a citizen working there, and the city can reposition its citizens to optimize the generated resources. Originally, the city is built with one citizen that works on the center tile. To acquire new citizens, the city must generate a surplus of natural resources. As the city's population grows, more tiles become engaged with work.
Each working tile generates a number of food, production and trade points per turn. Trade points can be turned into gold, luxury or science points. These six types of points -  \textbf{food, production, trade, gold, luxury and science} - constitute the  \textbf{city output}. 
In this way, we calculate the city output as a sum of all points that are collected with every turn, and doubling the production points as they can be used as half of a gold point when buying the current city project. The formula for the \textbf{city output} is given in equation \ref{eq:1}:
\begin{equation} \label{eq:1}
OUTPUT_T = \sum_{t=1}^{T}( gold_t + luxury_t + \\
science_t + food_t + production_t*2 + trade_t )
\end{equation}
where $t$ and $T$ refer to the turn number, and $t=1$ is the first turn of the game. For the cities built in later turns, all points prior to the turn in which they were built were taken as zeros.

Consequently, the \textbf{total game output} at turn $T$ is the sum of all city outputs owned by the player until the $T$-th turn:
\begin{equation}
TGO = \sum_{n=1}^{N} OUTPUT_{n,T}
\end{equation}
where $N$ is the number of cities owned by the player.

The complexity of settling in FreeCiv lies in the abundance of options to exploit the map tiles for natural resources and in the constrained nature of the player's possibilities. In general, the number of working tiles is in inverse ratio to the number of cities because every new settler unit removes one citizen from the home city. Consequently, home cities produce less resources and delay exploiting more working tiles. 
Therefore, maximal accumulated natural resources do not necessarily arrive form maximizing the number of cities. 
Moreover, placing the available citizens on different tiles within the city borders results in different amounts for each particular type, as well as cumulative, natural resources. Also, the type of the natural resources plays an important role in the city's development and, consequently, influences the total game output.  

We saw the estimation of the total game output as a regression problem that determines the relationships between the parameters of the map cluster and the TGO value. In other words, given a map cluster, we aimed to predict a continuous integer value reflecting the future contribution of the city being built in the cluster center to the TGO.

In FreeCiv, the amount of generated natural resources of each tile is affected by the map parameters, such as terrain type, the presence of special resources, rivers and improvements; and by the city economy, such as special buildings, city governor, the government type and trade routes.
Since we were mainly interested in the relationships between the map parameters and the city output, we only considered the parameters relevant to the map qualities. They are listed as following:
\begin{inparaitem}[\textbullet]
\item (TERR) Terrain of the center tile and terrain of the surrounding tiles within the map cluster. There are 9 possible terrain types in the game suitable for building a city: Desert, Forest, Grassland, Hills, Jungle, Mountains, Plains, Swamp and Tundra. 
\item (RES) Resources on the tile and surrounding tiles within the map cluster. Every type of terrain has a chance of an additional special resource that boosts one or two of the products.  Special resources can be one of 17 types and only one per tile.
\item (WATER) Availability of water resources. Presence of Ocean or Deep Ocean terrain in the city has special significance due to their rich resources and strategic advantages. Therefore, we considered them as extra parameters.
\item (RIVERS) Availability of rivers. Rivers enable improvements of the terrain and enhance trade for certain terrain types.
\end{inparaitem}

The outlined parameter list may not be exhaustive, however, we aimed to include the most relevant features that have the highest correlation with the generated natural resources.
The only two attributes unrelated to the map qualities were considered as those that characterize neighboring cities: number of player cities in the neighborhood (the 2 tiles wide area behind the city border) and the number of enemy cities in this region. We marked them 'MY\_NEIGHB' and 'ENEMY\_NEIGHB' respectively. 

With all parameters considered, the regression problem was set in the form shown in equation \ref{regression}:
\begin{equation} \label{regression}
f: (TERR, RES, WATER, RIVERS, \\
NEIGHB) \rightarrow TGO
\end{equation}
Equation \ref{regression} was addressed by two approaches: KB-RL and NN. The next sections provide detailed descriptions of both solutions.

It is possible that there are exiting solutions for the discussed task that are better than the two methods considered in this report. For example, such solutions may entail but are not limited to rooting in combinatorics, logic, or other areas of computer sciences. However, we did not aim to find the best solution, but rather saw this problem as suitable for comparing KB-RL with NN. Hence, it is possible that methods not addressed in this report can achieve better results when applied to the discussed problem.

\section{KB-RL METHOD} \label{KBRLMETHOD}

The KB-RL approach of the reported experiment is explained in detail in \cite{firstPaper}. Due to the restrictive nature of the conference paper, we provide here only a short summary of the method. 
The core of KB-RL is a knowledge-based system. KBSs were intensively studied in the 80s and 90s of the last century. Later, their popularity in the scientific community declined. However, there are numerous practical applications of the KBSs, such as \cite{KBSrevevaluation,oravec,empiricalstudy}. Also, there is much literature available on KBSs, for example \cite{AkerkarKBSbook}.
KBS encodes human knowledge into machine-readable rules for automated problem solving. A KBS typically consists of a knowledge base that holds the knowledge, and an inference engine that searches the knowledge to derive a valid conclusion for a given task. The inference engine in KB-RL processes the knowledge by adopting abductive reasoning. Abductive reasoning is characterized by the consideration of  contextual knowledge in the process of inferring a solution \cite{expertsystemsbook}. The knowledge base is represented by a semantic network that follows an ontology to define the formal specification of the used concepts and their relationships. KB-RL distinguishes between two types of knowledge: contextual knowledge and procedural knowledge. Contextual knowledge consists of two types: facts about the environment and its state, and the situational information (working memory). The procedural knowledge is encoded in the form of rules dictating what actions should be accepted under the specified contextual conditions.

Unlike the classical KBSs, where rule conflicts need resolution, KB-RL is able to handle conflicting knowledge, either contradictory, or redundant. It is particularly relevant to multi-expert knowledge, where different experts can have different opinions and diverse expertise about the problem domain. KB-RL views conflicting knowledge as the variations of the solution and applies RL to handle the conflicts with the objective to master the optimal solution for a specified task or problem. In the situation of conflict, the inference engine acts as an agent in the RL setting, treating rules as actions and using the learned policy to decide on the one rule to be executed.
The KB-RL approach employs the on-policy model-free Monte Carlo method as a RL algorithm for conflict resolution. The Monte Carlo method is characterized by averaging the state-action values and utilising an $\epsilon$-greedy policy to ensure state space exploration \cite{monteCarlo}. 

KB-RL inherited its ability to explain the derived solution and the corresponding reasoning process from the traditional KBSs \cite{explainableAI1}.
KB-RL provides visibility and information about the decision making with several mechanisms. Firstly, the knowledge is represented in an explicit, human-readable format that can be viewed by users and knowledge engineers. Moreover, the meta-knowledge in the form of ontology provides a common vocabulary that carries the semantics of real-world concepts, facilitating the understanding of the problem context \cite{explanationKBS}. Furthermore, by tracing the executed rules in the history of the solution, one can understand the contextual conditions in which the rule was applied \cite{alltogether}.  

In the reported experiment, we reused the implementation of playing FreeCiv with the KB-RL method presented in \cite{firstPaper}, including the ontology for the semantic network, most of the knowledge base, and the clustering model for the state space segmentation. The ontology entities represented the main concepts of FreeCiv, such as a game, a player, a unit, a city, and a tile. The properties of these concepts were implemented as attributes of the corresponding entities, while the edges reflected the relationships between the concepts. The semantic network contained all the game data as instances of the outlined ontology types. The data was kept in constant sync with the FreeCiv's environment and, therefore, the inference engine could operate always up-to-date searching for matching rules. For the RL algorithm, the clustering model was employed to reduce the game state space to the finite number of states. The model was trained by the k-mean algorithm, particularly Lloyd's algorithm with a maximum of 300 iterations. The dataset was created based on 1100 game histories published at \cite{freecivGames1}. The 31 features of the dataset were selected based on their correlation with the won/lost outcome of these games. The features were such as the game score, population size, the number of learned technologies, and others. The collected dataset had 386895 entries. As a result of experimenting with hyperparameters, 185 clusters were defined to represent the game state space.  

In \cite{firstPaper}'s knowledge base, the tiles' quality for the city output was calculated by one rule summing the food, trade and shield points of all tiles within the map cluster. Considering the complex relationship between the parameters of the map cluster and the amount of the natural resources that a city could generate being built in the centre of this map cluster, \cite{firstPaper}'s calculation of the tile quality is a rather simplified estimation. For our experiment, we replaced this rule with two different approaches: one was the Neural Network model (section \ref{nn}), and the other one was using the new rules for the knowledge base encoding several players' strategies for the tile evaluation.

\begin{figure}[t!]
\centering
\begin{minipage}{.5\textwidth}
  \centering
  \includegraphics[width=\linewidth]{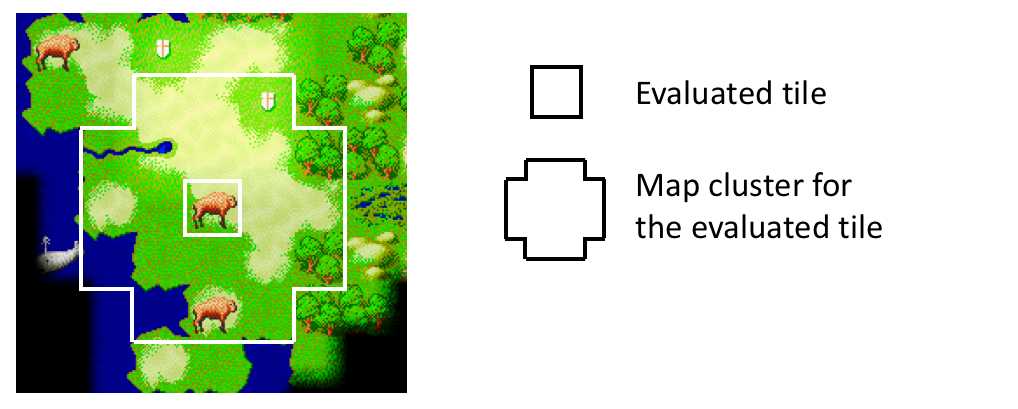}
  \caption{An example of a map cluster.}
  \label{fig:mapcluster}
\end{minipage}%
\begin{minipage}{.5\textwidth}
  \centering
  \includegraphics[width=\linewidth]{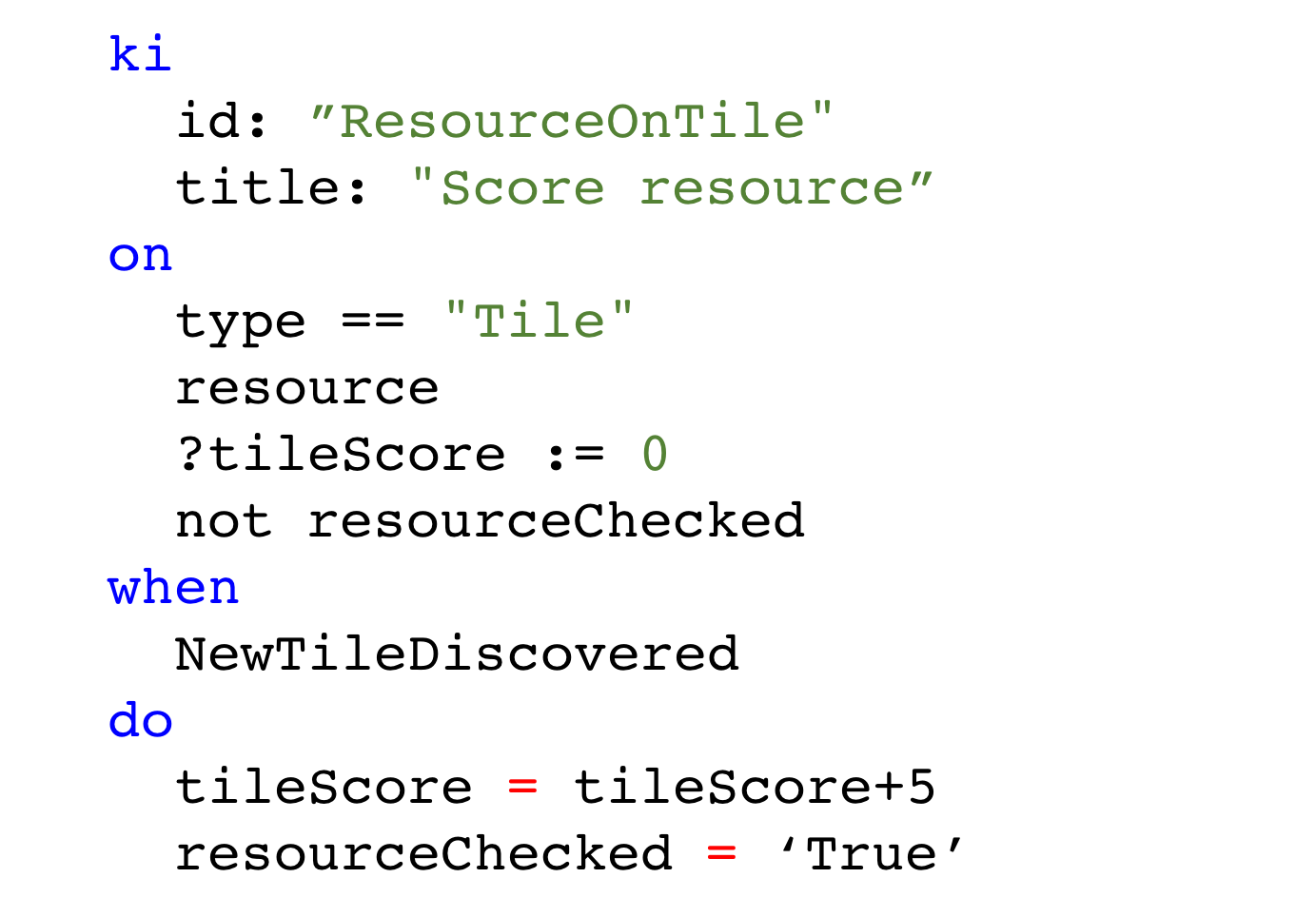}
  \caption{An example of a rule. The rule adds 5 points to the tile score for the special resource 'Bull' on the central tile.}
  \label{fig:ruleexample}
\end{minipage}
\end{figure}

Settling in FreeCiv has various strategies and can be solved by human players in numerous ways. Though the game rules are clearly defined, it takes practice for the players to master their settling strategies. Moreover, with more experience, players learn about different techniques that can be equally successful in winning the game. Following the KB-RL principle, we implemented several strategies into rules in order to evaluate the map clusters. These rules created a multi-expert knowledge base for the given task.  
A scoring system was used to estimate the tile quality to deliver high city output. Each tile held an attribute representing the tile quality score. The rules were encoded to manipulate the tile's score by adding more or less, or negative points to it based on the listed conditions. For example, the terrain of type Forest is highly valued by some players due to its high production output. Thus, the rule conditioning the Forest terrain added a high number of points to the tile score. Meanwhile, other players do not see this terrain type highly advantageous. Therefore, we added another rule that contributed to the tile score with less points. The terrain of type Desert earned negative reputation among players due to its scarce resources for city growth. These rules were adding negative or zero points to the tile score to reflect the rate of the players' dislike of the Desert tiles. Each rule contributed to the score of a particular tile independently of others, and rules conditioning the same features created the conflicting knowledge in terms of classical KBS. Following this principle, we could easily create more rules by adding various numbers of points for different map features to implement any player's strategy.
Figure \ref{fig:ruleexample} illustrates an example of a rule for the tile quality evaluation.

The rules addressed the parameter set outlined in section \ref{SelectedParameters}: terrain type, special resources and water resources on the central and surrounding tiles. There were 14 features covered by the rules: 9 for different terrain types and 5 for other features: (1) special resource on the central tile, (2) special resources on the surrounding tiles, (3) availability of water resources, (4) access to the deep ocean and (5) presence of the whale resource. As whales are resources that boost two products (food and production) at the same time, many players favor them over other resources. Thus, we treated it with additional rules. 

To cover different strategies, alternative rules were implemented for each feature adding a different amount of points to the tile score. For example, for the rule given in Figure \ref{fig:ruleexample}, the alternative rules added 1, 5, or 10 points to the attribute 'tileScore'. These numbers reflected the value that different human players would put, given the occurrence of a special resource on the tile when playing the game with different strategies. In the classical KBS, these rules would create a conflict due to the identical specified conditions. However, in KB-RL, the system had to learn by means of RL what rule would convey the best outcome for the designed goal.

Overall, 56 rules were written for the aforementioned 14 features. The created rules were added to the multi-expert knowledge base of report \cite{firstPaper}. It has to be noted, though, that the number of rules is rather indicative and does not reflect the complexity of the knowledge base.

For the purpose of this experiment, we had to redefine the reward function to reflect the connection between the total game output and the desired goal. One game was considered an episode and the total game output was calculated at the end of each game run.
Then, for every state $s$ of the episode, the reward $R_s$ was given in the amount of the total game output of this episode. Accordingly, the value of a state was calculated as the average of the acquired rewards: $V_s = avg(R_s(n))$ where $n \in N$, and $N$ is the set of all games that visited the state $s$. And, the state-action values were calculated by averaging the set of games $M$ that visited the state $s$ and performed action $a$: $q_s^a = avg(R_{s,a}(m)), m \in M$.

\section{NEURAL NETWORKS METHOD}\label{nn} 
Nowadays, Neural Networks do not need much of an introduction. A detailed discussion of Artificial Neural Networks can be found in \cite{NNbook}, or \cite{DeepLearning}.
NNs learn to perform a task by analysing the data rather than being explicitly programmed. Therefore, the data is a fundamental component for the NNs. Hence, collecting a dataset to train the NN model was our first task for implementing the NN solution in this experiment. For this purpose, we exploited 1100 fully played FreeCiv games from \cite{freecivGames1} that were played with the KB-RL approach reported in \cite{firstPaper}. The games were analysed, and all city places were identified as well as city outputs and total game outputs for all games. Based on the acquired information, the dataset was constructed for training NN to predict the quality of the place for the city output.

For each city place, the map cluster around the city was taken to create the input entry to the dataset, with the map cluster being centered on the tile where city was built. Firstly, the input data went through preprocessing - a common practice in machine learning engineering that usually includes procedures such as transforming data into vectors, normalisation and dealing with noise or missing values. In FreeCiv, the map is saved to a text file for each game turn being encoded into special symbols. These symbols represent all tile properties, such as terrain type, special resources, rivers, roads, tile owner and affiliation to the city. Based on these symbols, the map image can be entirely reconstructed at any point of time. Therefore, it was beneficial for us to use encoding, allowing us to convert the map cluster image into a feature vector with no loss of data integrity. Accordingly, the feature vector of the dataset included the following symbols: 21 symbols that reflect the terrain type of each tile in the map cluster, 21 symbols to indicate the presence/absence of the special resources on each of the 21 tiles and one symbol indicating the presence of the river on the central tile. Additionally, we counted the number of the player's cities and the number of the opponents' cities already built in the neighbourhood of the given place. We found it beneficial to ignore the position of the tiles within the city borders. The final dataset entry was constructed as columns where each column represented a symbol type, and the value was equal to the number of such symbols in the map cluster. For more detail, the dataset can be found at \cite{freecivGames}.

As a label column for the dataset entries, we estimated the output of the cities built on the derived map clusters. To do so, we faced a few challenges. Firstly, cities built on the same land in different games would differ in their output due to the different game development and the player's progress. Secondly, cities were built in various turns, but we had to estimate each city place independently from the turn built. Therefore, we could not use equation \ref{eq:1} to calculate the label for our dataset entries. By analyzing the data and experimenting with hyperparameters for training the neural network, we chose to calculate the city output as in equation \ref{eq:5}:
\begin{equation}\label{eq:5}
OUTPUT_c = \sum_{i=1}^{100} (gold_i + luxury_i + science_i \\
+ food_i + production_i*2 + trade_i)
\end{equation}
where c refers to the city index, and $i$ represents the age of the city in terms of turns. For example, $i=1$ relates to the first turn after the city was built, and $i=100$ is the 100th turn of city existence on the map. 

As a result, the corresponding dataset had duplicate entries with different labels for them. It was due to cities being built on the same spot in different games, or the parameters of the map clusters were the same but in different locations. We replaced the duplicate entries with the average of their label values.
By keeping only unique entries, we aimed to minimize potential data imbalance \cite{dataduplication}.

Accordingly, the resulting dataset contained 2765 unique entries for training the NN model. The input dataset was normalized by a min-max algorithm, and the trained model had the following structure:
\begin{inparaitem}[\textbullet]
\item Input layer accepts 83-dimensional feature vector.
\item       
One hidden layer with 95 neurons and ReLU activation.
\item       
Weights are initialized using normal distribution with zero mean and 0.0005 standard deviation.
\item       
To avoid overfitting, a dropout with probability 0.5 is applied to the hidden layer.
\item       
The output is a single neuron, which is a continuous variable.
\item       
The mean squared error is used as a loss function.
\item       
The ADAM optimizer is employed as it has shown the best performance among other optimization algorithms.
\item       
Batch size is 30 and learning rate is 0.002.
\end{inparaitem}

\begin{figure}[t!]
    \centering
    \includegraphics[width=0.8\linewidth]{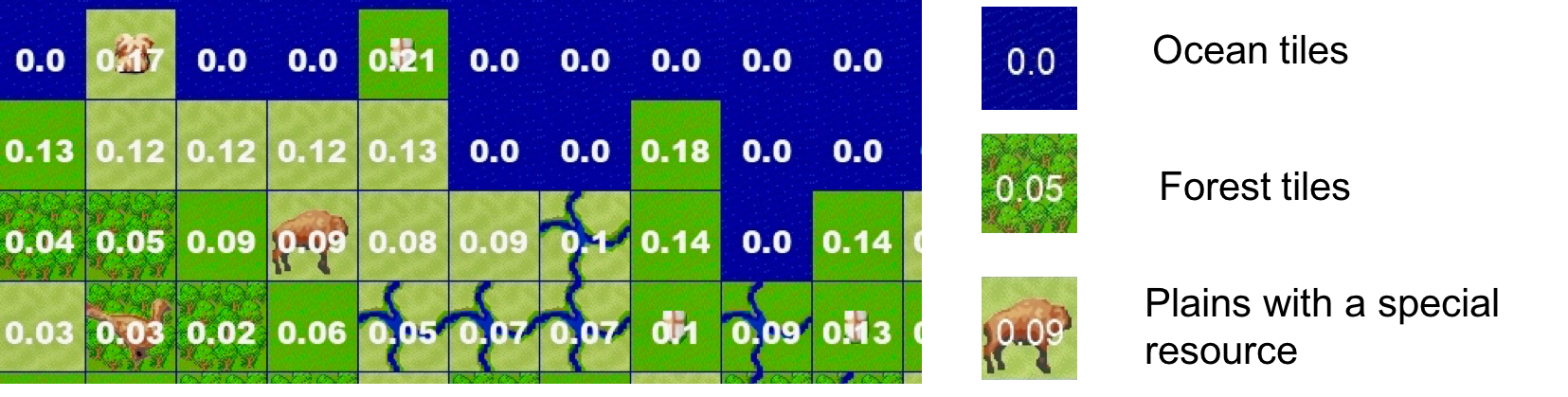}
    \caption{A fragment of the visual validation of the trained NN model. The number in the square is the score predicted by the NN model for an assigned tile.}
    \label{fig:modelvisual}
\end{figure}

In order to find optimal hyperparameters, including the number of hidden layers, grid search has been applied to the model. For the model assessment, we chose K-fold cross validation with 10 splits and with shuffling. After training, the mean squared error for the test set reached the value 0,00637. To our surprise, the grid search resulted in a single-layered model to be the optimal for the provided dataset. After careful consideration, we suggest that it can be due to a theoretically proved finding that a neural network with one hidden layer can adequately approximate "any Borel measurable function from one finite-dimensional space to another", given the sufficient number of hidden units \cite{reed1999neural,Goodfellow-et-al-2016}. To validate the trained model, we also performed visual inspection of the predicted tile scores for the different map fragments. Figure \ref{fig:modelvisual} shows that the model performed reasonably well, predicting higher scores for the tiles of Grassland and Plains terrain, especially if they were located on the coast or had a special resource.

The trained model was plugged into the KB-RL system in such a way that on the start of every game turn, the game map was preprocessed, the feature vector for every map cluster was created and passed to the NN for prediction. The predicted value was assigned as the tile's score representing the tiles' quality for the TGO.

\begin{figure}[t!]
\centering
\begin{minipage}{.5\textwidth}
  \centering
  \includegraphics[width=\linewidth]{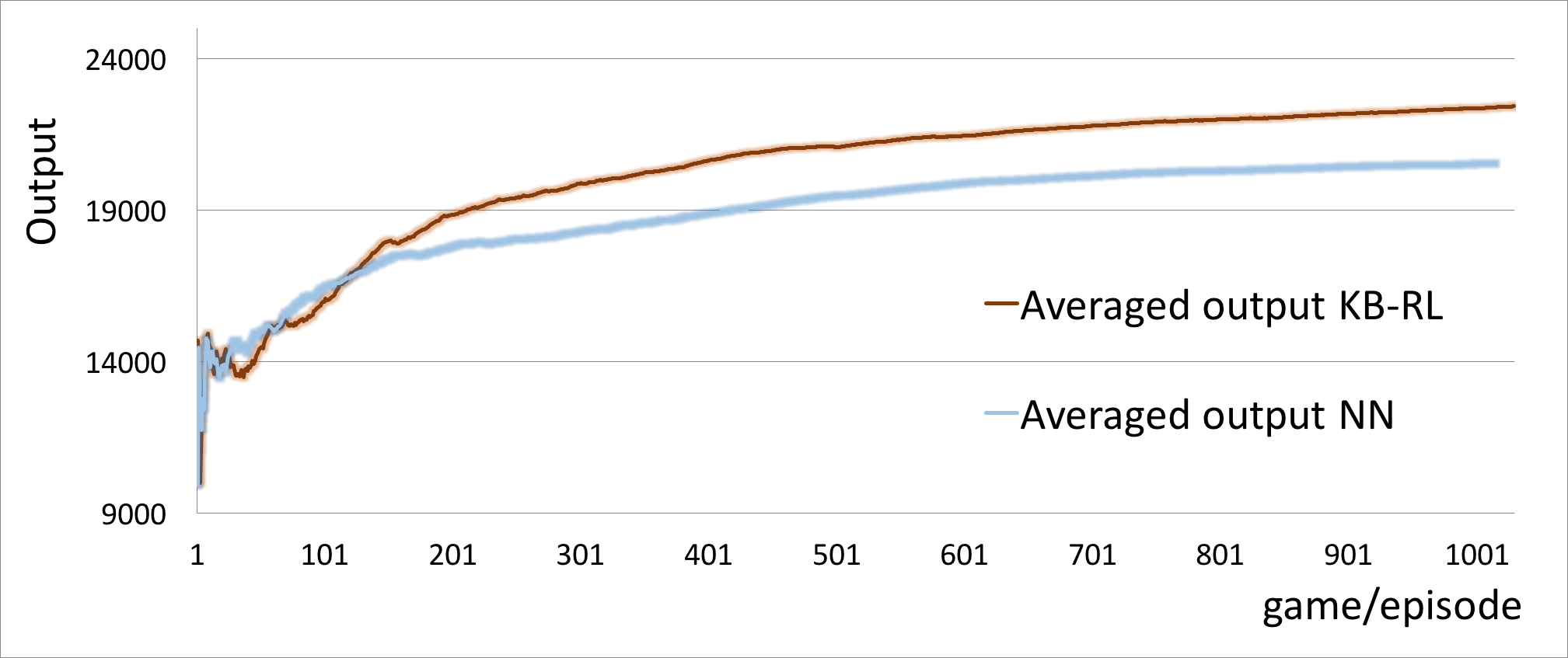}
  \caption{TGO averaged over a number of games.}
  \label{fig:outputchart}
\end{minipage}%
\begin{minipage}{.5\textwidth}
  \centering
  \includegraphics[width=\linewidth]{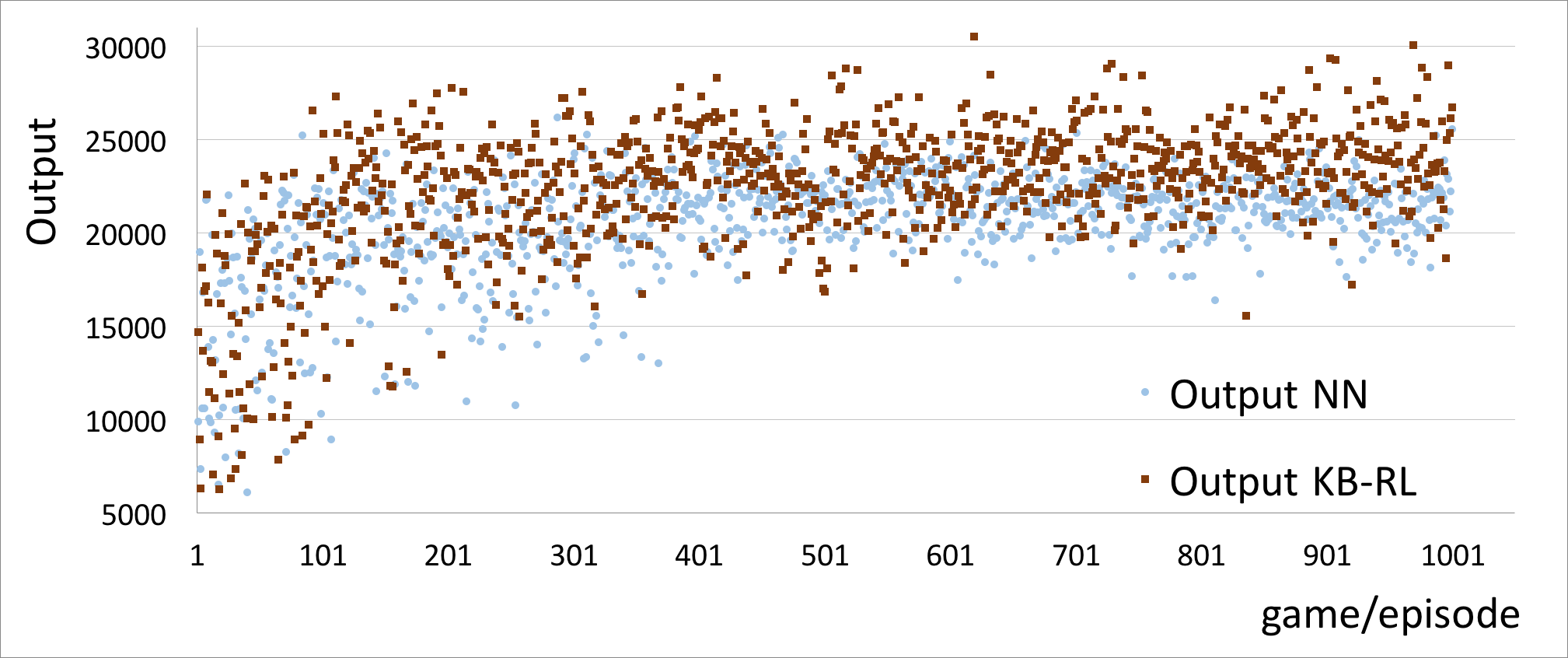}
  \caption{TGO for each single game.}
  \label{fig:outputchart1}
\end{minipage}
\end{figure}

\section{EXPERIMENTAL SETUP} \label{ExperimentSetup}

The experiment was set up in two ways. One time, the training of the KB-RL agent was conducted with the KB-RL method holding created rules for the evaluation of map clusters. The other time, the training ran with the plugged in NN model that replaced the rules for the evaluation of map clusters. For both runs, the TGO was observed and compared to estimate the performance of the two approaches.
This way, the difference in output of the two setups would result from the two approaches for the tile evaluation and, thus, would become a point of comparison for these two methods.

There were several design decisions to be made. Firstly, we had to decide on a starting point for the games. FreeCiv offers a game configuration either with a random starting point or with a predefined starting position. A random starting position would give more generalisation to the trained model and more balanced data. However, it would also bring a lot of randomness to the progress of the game. The starting position defines the player's nation, and the nation dictates many aspects of the game, such as the nation's temperament, the way it is treated by other nations and the citizens' loyalty to the leader. Therefore, the nation also influences the generated natural resources in the game. It would take a longer training time for the model to learn to recognize the effect of the random start versus the influence of the map parameters. On the other hand, the games used for creation of the Neural Network dataset were started at a fixed position on each of 5 maps \cite{firstPaper}. Based on this consideration, the decision was made to start the games from the fixed staring position.  Given that the conditions for both setups were equal, we believe that this decision did not diminish the outcome of the experiment. 

For the game configuration, we chose one of the configurations described in \cite{firstPaper} named Chaos. The map of this configuration is 80x50 tiles with two big continents in the middle, and five small islands on the south of the map. The configuration is designed to be played by two KB-RL agents, where each of them starts the game on the separate continent. No embedded AI players are added to the game, however, Barbarians are activated. This configuration favours the game conditions where the players have no contact in the initial game phase to be engaged in warfare or diplomacy, and, therefore, are encouraged to build and develop cities. 

Another design decision was made about the length of the games.
Settling happens in FreeCiv in the initial phase. After cities are built, the player mostly focuses on developing the economy, technologies and warfare. For the purpose of the experiment, we did not need to play the games until they are finished. Stopping the games ahead of time gave us the advantage of significantly shorter episode duration: such episodes took about $10\%-20\%$ of the full game time. Analyzing the histories of games from \cite{firstPaper}, we chose to play only the first 120 turns of the game, as it seemed to be a good trade-off among the amount of generated data, the game state and the play time.

Initially, we also considered the setup where learning would be performed only once, with added rules for tile evaluation. Then, the performance would be observed for the total game output with the learned policy and without ongoing learning. After that, the rules would be replaced by the NN model, and the system performance would be measured again. However, in this case the NN might lead the game through a different set of states that were not well-learned during initial training. This fact would be a disadvantage to the NN setup. Thus, the decision was made in favor of running learning for both setups from the same starting point.

\section{RESULTS}

\begin{figure}[t!]
\centering
\begin{minipage}{.5\textwidth}
  \centering
  \includegraphics[width=\linewidth]{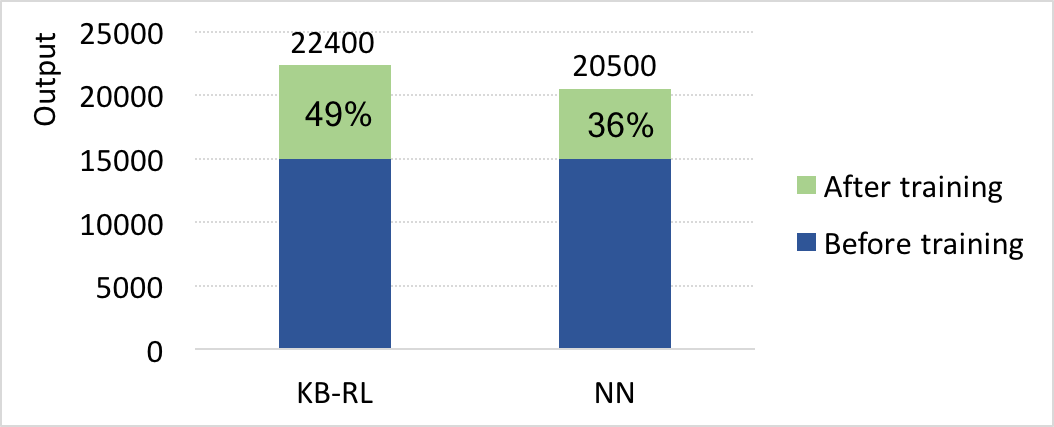}
  \caption{TGO improvement after training for the KB-RL and the NN setups.}
  \label{fig:increase}
\end{minipage}%
\begin{minipage}{.5\textwidth}
  \centering
  \includegraphics[width=\linewidth]{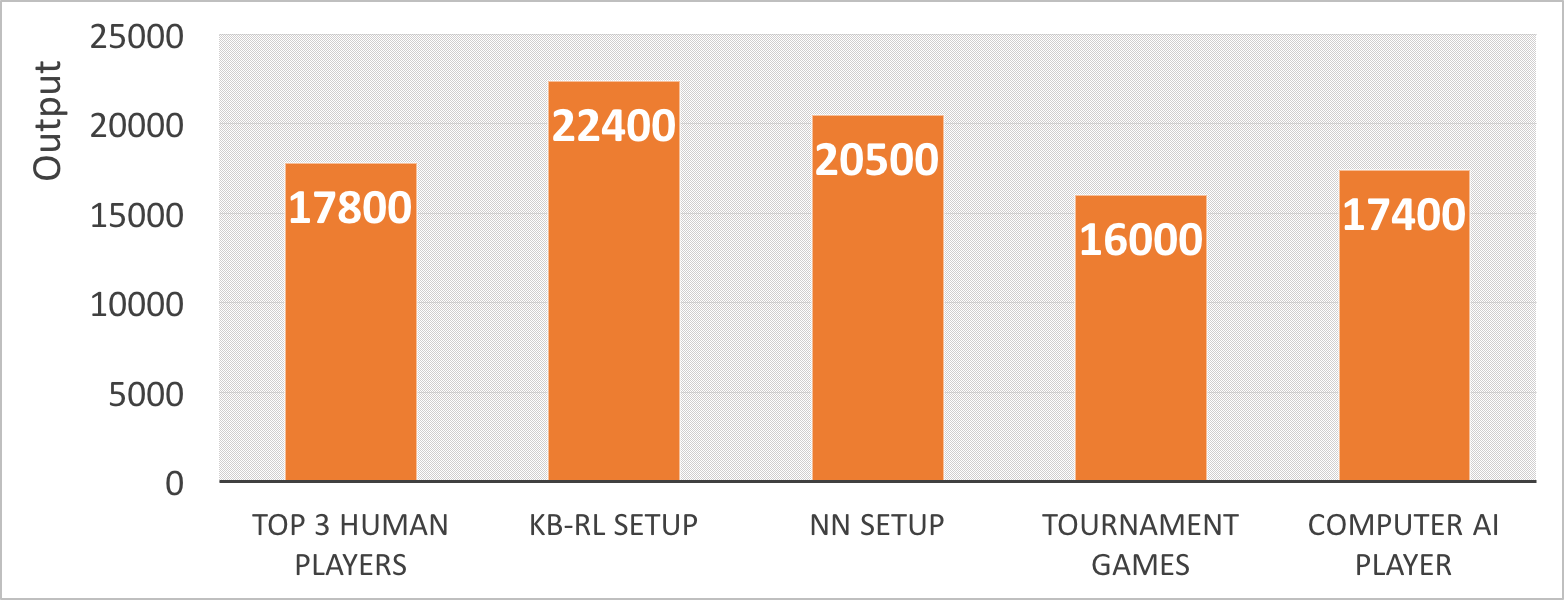}
  \caption{Average TGO for the KB-RL and the NN setups in contrast to the human players, the tournament games and the computer AI players.}
  \label{fig:compare}
\end{minipage}
\end{figure}

To measure the performance of both experiments, we chose the metric of averaging the total game output over the number of games. The averaged total game output was calculated after every game and observed for both setups in comparison to each other. Figure \ref{fig:outputchart} visualizes the averaged TGO progress throughout the training for the two setups. Additionally, Figure \ref{fig:outputchart1} shows the TGO of each single game in the run of the episodes. At the beginning, the TGOs had much variation, with the average value just under 15 000. As learning proceeded, the TGO steadily climbed up, and the variation declined. This behaviour was common for both experiments as a result of the reinforcement learning optimization within the KB-RL system.

Each experiment ran for 1000 episodes. At the end, the game play stabilized with an average total reward of 20 500 and 22 400 points for the NN and the KB-RL setups respectively. 
For the KB-RL setup, the improvement constituted $49\%$ in contrast to the starting value, while for the NN, it was a $36\%$ increase ( Figure \ref{fig:increase}). The difference of $13\%$ between the two approaches stems from the different solutions for the tile evaluation. 

To better understand the achieved results, we compared them to the performance of the human players, FreeCiv's computer AI players and the so-called tournament games of report \cite{firstPaper}. Figure \ref{fig:compare} illustrates the compared TGOs. The games played by humans were acquired as a part of the reused data of the report \cite{firstPaper}, they are published online at \cite{freecivGames}. For comparison, we show the TGO of the top 3 players. Analysing the histories of their games, it can be concluded that they are definitely great experts in playing the game as their play was quick and efficient, and they won against computer AIs with a big winning margin. This comparison showed that both trained agents outperformed human and computer AI players (Figure \ref{fig:compare}). It must be noted, however, that neither human, nor computer AI players explicitly pursued the goal of maximizing generated natural resources. Therefore, this comparison is rather indicative.  

\begin{figure}[t!]
\centering
\hspace{5mm}
\begin{minipage}{.4\textwidth}
  \centering
  \includegraphics[width=\linewidth]{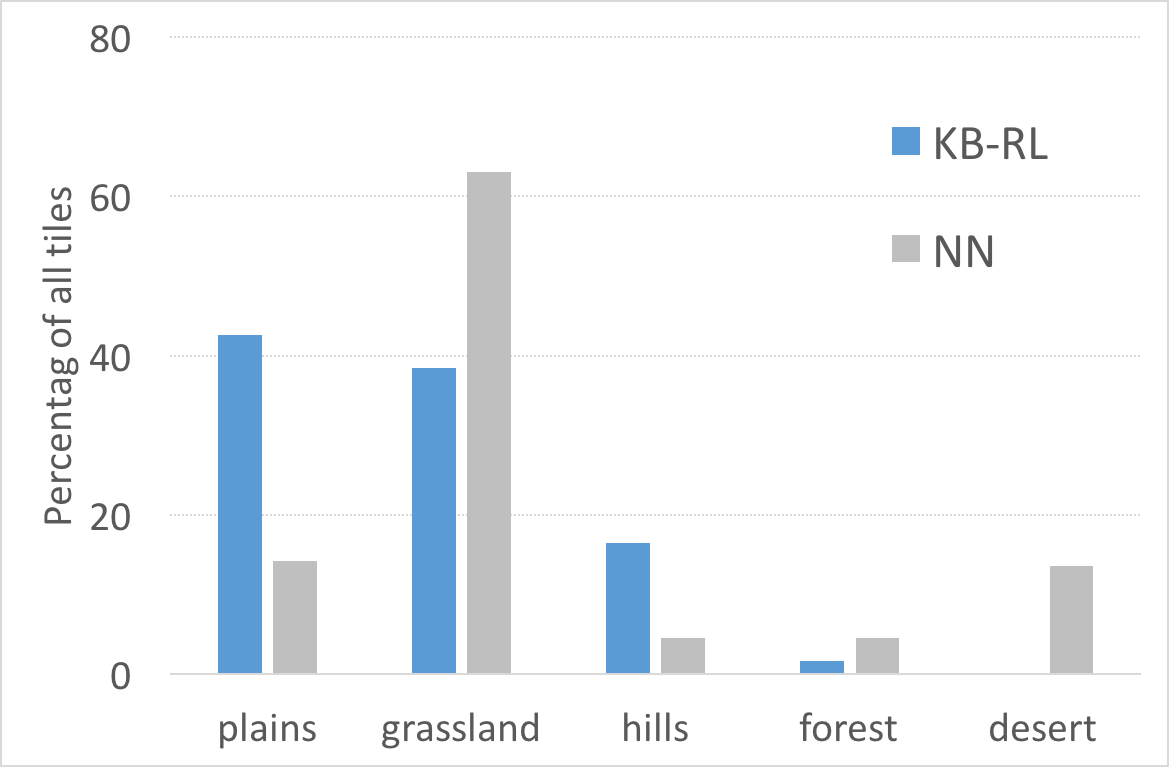}
  \caption{Distribution of the terrain types on the city's central tile.}
  \label{fig:main_terr}
\end{minipage}%
\hspace{5mm}
\begin{minipage}{.5\textwidth}
  \centering
  \includegraphics[width=\linewidth]{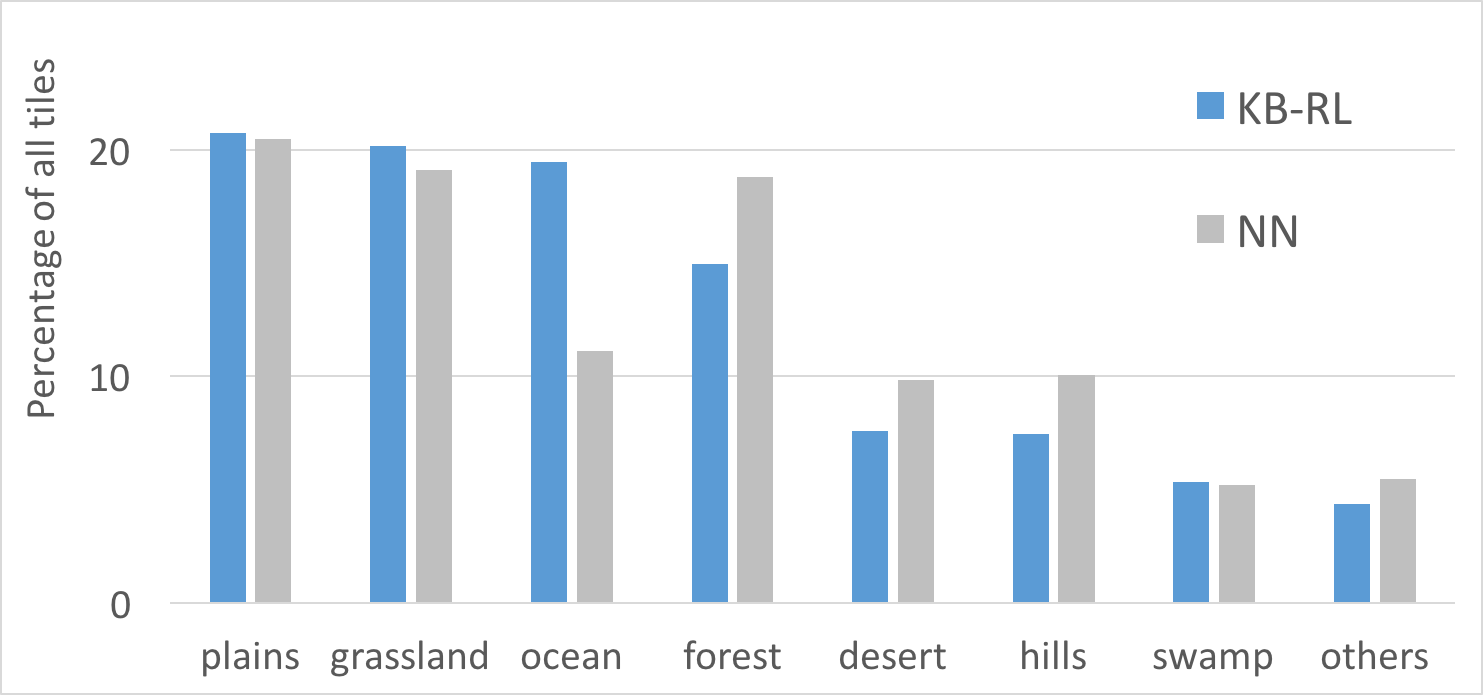}
  \caption{Distribution of the terrain by type after training.}
  \label{fig:terr_around}
\end{minipage}
\end{figure}

When investigating the two setups in contrast to each other, it can be seen that the fundamental difference in building cities was in the terrain type of the central tile. Furthermore, there was a noticeable difference in the terrain type of the tiles surrounding the central tile within city borders. While both setups built comparatively similar number of cities, with the similar amount of resources and rivers within city borders, the terrain type of the city tiles differs significantly (Figures \ref{fig:main_terr} and \ref{fig:terr_around}).
In the KB-RL setup, the majority of cities were built on one of the three terrain types: plains ($42\%$), grassland ($38\%$) and hills ($16\%$). On the contrary, most of the cities in the NN setup were built on grassland ($63\%$) with a surprisingly big part of cities being built on the desert terrain ($13\%$). Most likely, this is a consequence of a data deficit during the training of the NN model as the desert terrain is an obvious disadvantage for the city's development. 

Cities of both setups occupied the terrain of types grassland and plains to a similar extent (Figure \ref{fig:terr_around}). However, the KB-RL agent tended to build cities mostly on the coast, occupying a high number of ocean tiles. At the same time, the NN setup showed more preference towards the forest terrain, while coastal terrains were occupied considerably less than forest. Furthermore, cities in the NN setup occupied more terrain of types hills and desert in comparison to the KB-RL setup.

While explaining the NN's prediction for the tile scores in the NN setup was challenging for us, the games' execution histories of the KB-RL setup gave us a valuable insight into the decision-making of the KB-RL agent. By reviewing the execution histories, we were able to check which rules contributed to the score of every tile. As the rules were encoded in the human-readable format using the semantically loaded vocabulary, it was straightforward to understand the reasoning process of the KB-RL agent. For example, for the map cluster in Figure \ref{fig:mapcluster}, the executed rules were ResourceOnTile (added points for special resource bull on the central tile), ResourcesAround (added points for the two special resources within the map cluster), TerrainGrassland (added points for the Grassland terrain type of the central tile) and OceanTileBonus (added points for the Ocean terrain type of two tiles within the map cluster). Altogether, they added 27 points to the tile score. This number varied from game to game depending on the conflict resolution outcome since each of these rules had a set of the alternative rules, as it was explained in section \ref{KBRLMETHOD}. For each rule, the execution history listed its conflicting rules and their probabilities at the time of conflict resolution. We saw a strong possibility to use this information in the analysis of the training process and its progress. However, this task was not included into this report and, potentially, could become a research study on its own.

\section{CONCLUSION}

The goal of this article was to study the KB-RL approach by comparing its performance to the NN in solving a classical Artificial Intelligence task. The evaluation of map tiles for building cities to deliver high output of the generated natural resources was chosen for this comparison.
The results of the experiment support the idea that KB-RL can be a worthy alternative to other AI approaches, such as Neural Networks, in solving tasks of the Artificial Intelligence field.

By leveraging human experts' knowledge, KB-RL performed the initial phase of the training equally well as the NN, which was trained on the 1100 previously played games. These results advocate that KB-RL may be a better choice for the problems where the data for training a NN model is scarce or unavailable, but there is an access to human expertise.
Moreover, with more games played, KB-RL demonstrated a greater ability to optimize the complex policy, justifying the smaller demand of the data by KB-RL to deliver a better policy.

Another important advantage of the KB-RL solution was the ability to explain the system decisions by reviewing the rules executed in the game. 
This fact can be imperative in cases where the system's explainability is a critical requirement.

\bibliography{main}
\end{document}